\documentclass[11pt, a4paper, logo, twocolumn, copyright]{googledeepmind}
\usepackage[authoryear, sort&compress, round]{natbib}
\usepackage{fvextra}
\usepackage{hyperref}
\usepackage{graphicx}
\usepackage{siunitx}

\usepackage{listings}
\title{Vision-Language Model Dialog Games for Self-Improvement}

\correspondingauthor{kksenia@google.com}


\author[1]{Ksenia Konyushkova}
\author[1]{Christos Kaplanis}
\author[1]{Serkan Cabi}
\author[1]{Misha Denil}

\affil[1]{Google DeepMind}

\begin{abstract}
The increasing demand for high-quality, diverse training data poses a significant bottleneck in advancing vision-language models (VLMs).
This paper presents VLM Dialog Games, a novel and scalable self-improvement framework for VLMs. 
Our approach leverages self-play between two agents engaged in a goal-oriented play centered around image identification.
By filtering for successful game interactions, we automatically curate a high-quality dataset of interleaved images and text.
We demonstrate that fine-tuning on this synthetic data leads to performance gains on downstream tasks and generalises across datasets.
Moreover, as the improvements in the model lead to better game play, this procedure can be applied iteratively.
This work paves the way for self-improving VLMs, with potential applications in various real-world scenarios especially when the high-quality multimodal data is scarce.
\end{abstract}

\begin{document}

\maketitle

\section{Introduction}
Large language models (LLMs) have achieved remarkable success by training on vast datasets that now include a significant portion of the Internet~\citep{achiam2023gpt, gemini2024}. 
Their performance generally scales with training data size~\citep{kaplan2020scaling}, but acquiring new, high-quality data is increasingly challenging, especially for vision-language models (VLMs), which require carefully curated interleaved image and text data.
Recent research~\citep{chen2024selfplay,bai2022constitutional, yuan2024selfrewarding, huang2022large} indicates that self-improvement techniques can use synthetically generated data to overcome this limitation.
We introduce a novel self-improvement method based on goal-oriented play between VLMs.
This approach provides a scalable way to iteratively generate high-quality synthetic data, which can be used to fine-tune the model for further performance improvement.
By carefully designing the game, we can target specific capabilities and domains for improvement, while the goal-oriented nature ensures the quality of the generated data.

\begin{figure}[h]
    \centering
    \includegraphics[width=\columnwidth]{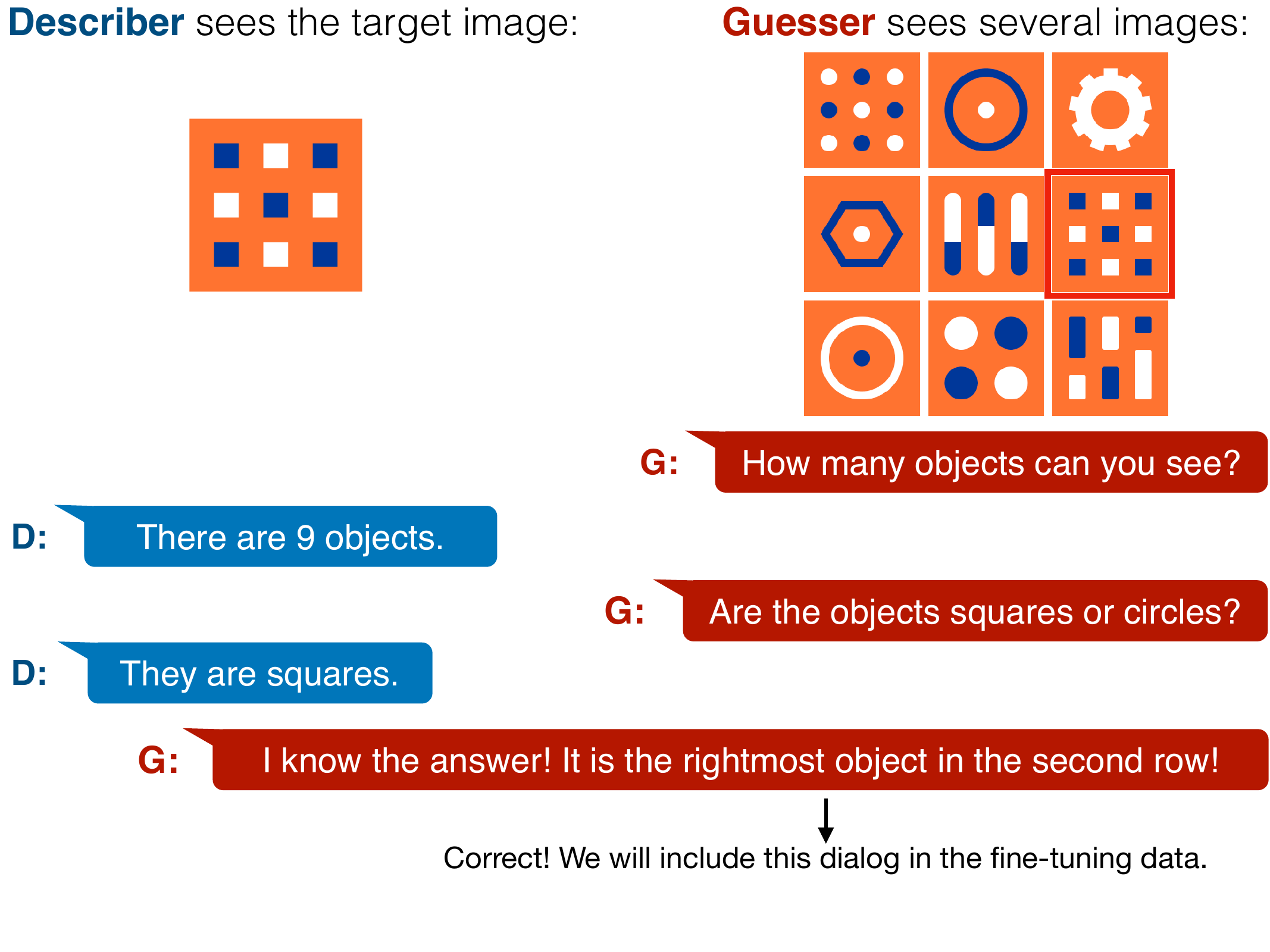}
    \vspace{-8mm}
    \caption{\textbf{Example interaction between a Guesser and a Describer in the VLM Dialog Game.} The Guesser aims to identify the target image from a set of distractors by asking questions, which the Describer answers. Since the Guesser correctly identifies the target image at the end of the game, this dialog is considered successful and included in the fine-tuning data.}
    \vspace{-6mm}
    \label{fig:game-example}
\end{figure}

We initiate the process with two VLMs which are assigned the roles of "Describer" and "Guesser" in a variant of reference game~\citep{krauss1964changes, das2017visual, de2017guesswhat, hakimov2024usinggameplayinvestigate} which we call "VLM Dialog Game".
Using a set of unlabelled images, we construct a game with one target and several distractor images. 
The Describer answers questions about the target image, while the Guesser poses targeted questions to disambiguate the target from the distractors (Figure~\ref{fig:game-example}). 
While similar games exist, their primary use has been for human-based data collection or VLM evaluation, and our work demonstrates that this game framework also facilitates VLM iterative self-improvement through goal-oriented self-play.

Thanks to their instruction-following and image-understanding capabilities, the pre-trained VLMs achieve a non-zero success rate in this game.
This inherent ability provides a scalable method for generating interleaved image-text data.
The initial performance is imperfect:
The Describer may provide incorrect answers, and the Guesser may ask irrelevant questions.
However, the game's structure allows us to identify successful game instances where the Guesser correctly selects the target image.
By filtering for these successful dialogs, we automatically obtain a high-quality dataset of interleaved data.
This curated dataset is then used to fine-tune the VLM, improving its proficiency in playing the game and, consequently, its overall image understanding capabilities.
The improved model has a higher success rate in the game, and it can be further used to generate a better dataset of interleaved data, enabling further cycles of improvement.

Our experiments demonstrate that fine-tuning VLMs on the dialog game data yields significant improvements, not just in game performance, but also on related image understanding benchmarks.
Fine-tuning on games based on either OpenImages~\citep{kuznetsova2020openimages} or DOCCI~\citep{OnoeDocci2024} datasets leads to an increase in accuracy on VQAv2~\citep{goyal2017making} benchmark.
The approach is adaptable to specific domains, like robotics, where high-quality data is often scarce.
When the game is designed with frames from robotics episodes we significantly improve the model's ability to detect successful execution in robotics tasks.

The remainder of this paper is organized as follows: Section~\ref{sec:relwork} reviews related literature, Section~\ref{sec:method} describes the VLM dialog game and self-improvement methodology, Section~\ref{sec:exps} presents experimental results in general VQA and robotic success detection, and finally Section~\ref{sec:conclusions} summarizes findings, discusses limitations, and outlines future research.
\section{Related Work}
\label{sec:relwork}

\paragraph{Dialog games}

Various forms of dialog reference games have been known in linguistics since a long time~\citep{krauss1964changes}.
In computer science, prior work on multimodal dialog games is primarily focused on collecting datasets of grounded dialogs \citep{haber2019photobook, das2017visual, de2017guesswhat} or more recently, evaluating the capabilities of VLMs \citep{hakimov2024usinggameplayinvestigate, chalamalasetti2023clembench}.  
These existing games vary in design, including the number of images involved (single or multiple), the roles of the agents (symmetric, sharing the same goal, or asymmetric), and the interaction length (single-turn or multi-turn).

In contrast to these evaluation and data collection efforts, our work leverages dialog games for self-improvement, creating synthetic datasets to enhance VLM capabilities.
To the best of our knowledge, this is a novel use of the dialog games.

\paragraph{Self-Improvement}

Self-improvement~\citep{chen2024selfplay} techniques have gained significant interest in both language and multimodal learning.
A prevalent approach involves using an LLM to critique and refine its own responses \citep{bai2022constitutional, yuan2024selfrewarding}.
For instance, \citet{huang2022large} demonstrate that fine-tuning on self-generated rationale-augmented answers, without ground truth labels, enhances LLM reasoning. 
\citet{subramaniam2025multiagent} propose a multi-agent framework where diverse responses from a society of LLMs drive iterative fine-tuning and continuous improvement.

Self-improvement has also shown promise in enhancing multimodal understanding of VLMs, albeit with fewer existing studies.
A prominent technique in VLM self-improvement is cycle consistency, initially developed for image-to-image translation \citep{zhu2017cyclegan}.
It enforces that a transformation from a source domain to a target domain, and back, yields an output similar to the input.
This principle has been successfully extended to the multimodal domain \citep{li2023leveraging, li2023dalle, sharifzadeh2024synth}, often exploiting the symmetry between image captioning and text-to-image generation.
Cycles such as text1 $\rightarrow$ image  $\rightarrow$ text2 or image1  $\rightarrow$ text  $\rightarrow$ image2 aim to ensure similarity between the initial and final elements (text1 \& text2, or image1 \& image2) while producing data for self-improvement. 
Cycle consistency is particularly valuable when paired text-image data is limited \citep{li2023leveraging}, but can also facilitate the generation of novel image compositions with abundant data \citep{sharifzadeh2024synth}. 
Another approach to improving VMLs performance is through generating synthetic datasets by eliciting detailed question-answer datasets~\citep{luu2024questioning}.
Furthermore, self-improvement in VLMs is often tailored to specific applications, such as medical imaging, where data acquisition is challenging \citep{wang2024self}.

In contrast to many LLM self-improvement methods that rely on agent-based response generation and critique, we propose a novel self-improvement framework based on games. 
While inspired by the underlying principle of cycle consistency, our approach deviates from the traditional image-text-image cycle.
Instead of direct image generation, we map a target image to a dialog, and then back to the target image through contrastive image selection thus eliminating the need for text-to-image model.
\section{Method}
\label{sec:method}

This section introduces our method for iterative self-improvement through VLM Dialog Games.
We first describe the VLM Dialog Game mechanism and its key properties: self-play and goal-oriented nature, which are crucial for self-improvement (Section~\ref{method:game-desc}).
We then detail the complete self-improvement workflow, including game setup, dialog generation and filtering, model finetuning, final evaluation on the target task, and potentially repeating these steps (Section~\ref{method:workflow}).

\subsection{VLM Dialog Game Description}
\label{method:game-desc}

We introduce a VLM Dialog Game which is constructed using unlabelled images and two VLM agents.
The first agent, the \textbf{Describer}, is presented with a single target image and is instructed to faithfully answer questions about it. 
The second agent, the \textbf{Guesser}, receives a set of $N$ images, including the target image and several distractor images.
The \textbf{Guesser}'s objective is to identify the target image by posing questions to the \textbf{Describer}.
The agents' behaviour is controlled by prompting mechanism for VLMs which is described further in Section~\ref{method:impldet}.

Figure~\ref{fig:game-example} shows an illustrative example of the VLM Dialog Game in action. 
All images that the Guesser sees contain white and blue objects on an orange background, thus, to identify the target image the Guesser should focus on more specific properties of the images.
To disambiguate, the Guesser initiates a series of clarifying questions, such as "How many objects can you see?" and "Are the objects squares or circles?".
Once the responses uniquely define the target image (in this case, by pointing to "\num{9} square objects"), the Guesser successfully identifies it.
While resembling a classic reference game used for human data collection and VLM evaluation, this specific design features two key elements enabling VLM self-improvement: self-play for data generation and automatic success determination.

\paragraph{Self-play}
Current VLMs, thanks to their instruction-following capabilities, demonstrate a non-trivial success rate in playing this dialog game \citep{hakimov2024usinggameplayinvestigate}.
This capability enables a scalable approach to data collection through self-play of two prompted models.

\paragraph{Success Determination}
We use the Guesser's final selection to determine the success of the dialog: If the selected image matches the target image, the dialog is considered successful and added to the synthetic training data, otherwise it is discarded. 
This mechanism provides crucial, automatic quality control.

\subsection{Workflow}
\label{method:workflow}

The properties of the VLM Dialog Game enable the following workflow for self-improvement:

\begin{itemize}
    \item \textbf{Game setup}: Configure the dialog game with a designated unlabelled image dataset.
    \item \textbf{Dialog generation}: Generate dialogs via self-play between the VLM agents.
    \item \textbf{Dialog filtering}: Filter generated dialogs based on the success criteria.
    \item \textbf{Model improvement}: Fine-tune the VLM using the filtered dialog data and evaluate its performance on the target task.
    \item \textbf{Repeat the above steps (if needed)}: Repeat dataset generation with an improved version of VLM.
\end{itemize}

\subsubsection{Game Setup}
\label{method:impldet}

This section details the setup of the VLM Dialog Game, including agent instructions and image selection strategies.

\paragraph{Agent instructions}
We provide precise instructions to both the Describer and Guesser agents to guide their interaction in the game. 
The Describer is instructed to answer questions about the target image truthfully and accurately.
The Guesser agent operates in two stages:

\begin{enumerate}
    \item Questioning/Guessing: Initially provided with an empty image description, the Guesser must either:
    \begin{itemize}
        \item Ask a clarifying question to distinguish the target image from the distractors, or
        \item Make a guess identifying image X as the target image where X is the index of the hypothesized target image among the distractors.
    \end{itemize}
    \item Summarisation: The Guesser must create a concise summary of the target image description given the initial image description (or the previous summary), a question and an answer.
\end{enumerate}
Specific prompt texts for both agents are provided in Appendix~\ref{game-prompts}.

\paragraph{Image selection and game difficulty}
The images used in the game can be sourced from various datasets, including general datasets of natural images like OpenImages~\citep{kuznetsova2020openimages} or DOCCI~\citep{OnoeDocci2024}, or domain-specific datasets tailored to applications such as robotics~\citep{zhao2023learning}.
The game's difficulty is controlled through two primary factors related to which images are selected for a game:
\begin{itemize}
    \item Number of distractors: Increasing the number of distractor images directly increases difficulty. This is due to: (1) the Guesser needing to attend to a larger context, and (2) an increased likelihood of a distractor closely resembling the target image.
    \item Image similarity in each game: Randomly selecting images from the dataset creates an easier game, while grouping visually or semantically similar images increases the challenge.
\end{itemize}

We select the appropriate settings in the game so that the games are sufficiently difficult to produce interesting dialogs, but still feasible so that we generate sufficient amount of synthetic training data.

\subsubsection{Dialog Generation}
During this stage, the Describer and Guesser agents engage in an interactive dialog.
We construct the training dataset from examples of successful behavior by both the Describer and the Guesser:
\begin{itemize}
    \item \textbf{Describer examples.} Input: a single image and a question about it; Output: the corresponding answer.
    \item \textbf{Guesser examples.} Input: $N$ images (including the target and distractors) and a cumulative summary of the target image description; Output: either a clarifying question or a guess identifying the target image.
\end{itemize}
Each successful VLM Dialog Game generates multiple training examples of both types.

\subsubsection{Dialog Filtering}
\label{method-filtering}

The game's design allows us to directly verify the Guesser's final selection.
However, to mitigate the possibility of correct guess occurring by chance, we perform an additional validation step. 
We re-run the dialog without the final selection using the same images but in a permuted order and verify that the correct target image is identified in all cases.
Empirically, we observed that the position of the \textit{target} image has the most significant impact on the Guesser's accuracy, while the relative order of the distractors (given a fixed dialog) has a smaller effect.
Therefore, for computational efficiency we limit the tested permutations to $N$ where we ensure that the target image appears at each possible position (\num{1} to $N$), while the distractors order can remain fixed.
The datapoints from these consistently successful games form the filtered dataset for subsequent model training.

\subsubsection{Model Improvement}

The filtered dataset from successful dialog games is then used to fine-tune the VLM in a standard supervised fine-tuning way.
If the gains in playing the VML Dialog Games are large, we can use the improved model in order to collect the new synthetic dataset for further model improvement.
While this process directly affects the VLM's performance within the dialog game itself, our primary focus is on evaluating its capabilities on downstream tasks.
For instance, if the dialog game utilizes images from a robotics domain, we assess the fine-tuned VLM's performance on tasks such as robotic success detection, and for general images we test the performance on visual question answering (VQA) on the unseen images.
\section{Experiments}
\label{sec:exps}

\subsection{Experimental Setup}
\label{sec:exps-setup}

We evaluate our method using the Gemini 1.5 Flash model \citep{gemini2024} as the base VLM. 
Gemini 1.5 Flash is a powerful, instruction-tuned VLM that can take as input interleaved text and images and it provides a strong base model.
We use standard supervised fine-tuning procedure (see Appendix~\ref{sec:training-details}).
We limit the game length to a maximum of three question-answer turns.
For conciseness, we refer to the self-improvement method of the fine-tuning on synthetically collected dialogs as "VLM Dialog Games".

\subsection{Experiments with General Images in Dialog Games}
\label{sec:exps-docci}

This section details our experiments using the DOCCI~\citep{OnoeDocci2024} and the OpenImages datasets \citep{kuznetsova2020openimages} to evaluate the effectiveness of our self-improvement method for image understanding through VQA tasks.

\subsubsection{Dataset and Game Configuration}
\label{sec:game-config}

\begin{figure}[t]
    \centering
    \includegraphics[width=\columnwidth]{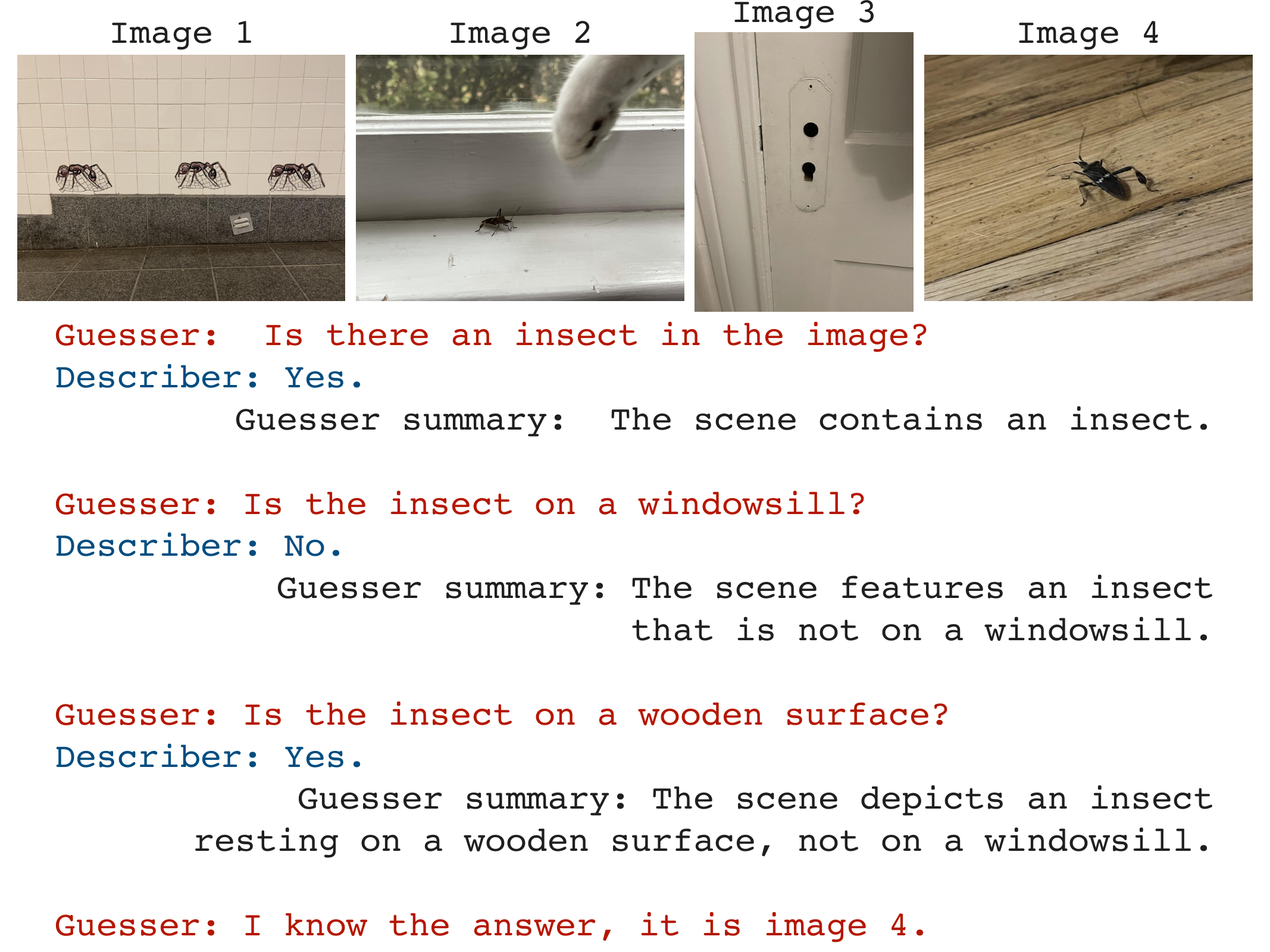}
    \caption{\textbf{An example dialog game using images from the DOCCI dataset}, grouped by clusters.
    The figure shows the Guesser's questions, the Describer's answers, and the Guesser's internal dialog summary.  The Guesser correctly identifies the target image (4) at the end of the dialog.}
    \vspace{-3mm}
    \label{fig:docci_example}
\end{figure}

\paragraph{DOCCI} dataset contains clusters of images grouped by their category.
We randomly sample \num{1000} image groups, each containing $N = 4$ images from one of \num{149} categories.
Figure~\ref{fig:docci_example} provides an example of a dialog game generated by prompted Gemini using this setup.

\begin{figure}[t]
    \centering
    \includegraphics[width=\columnwidth]{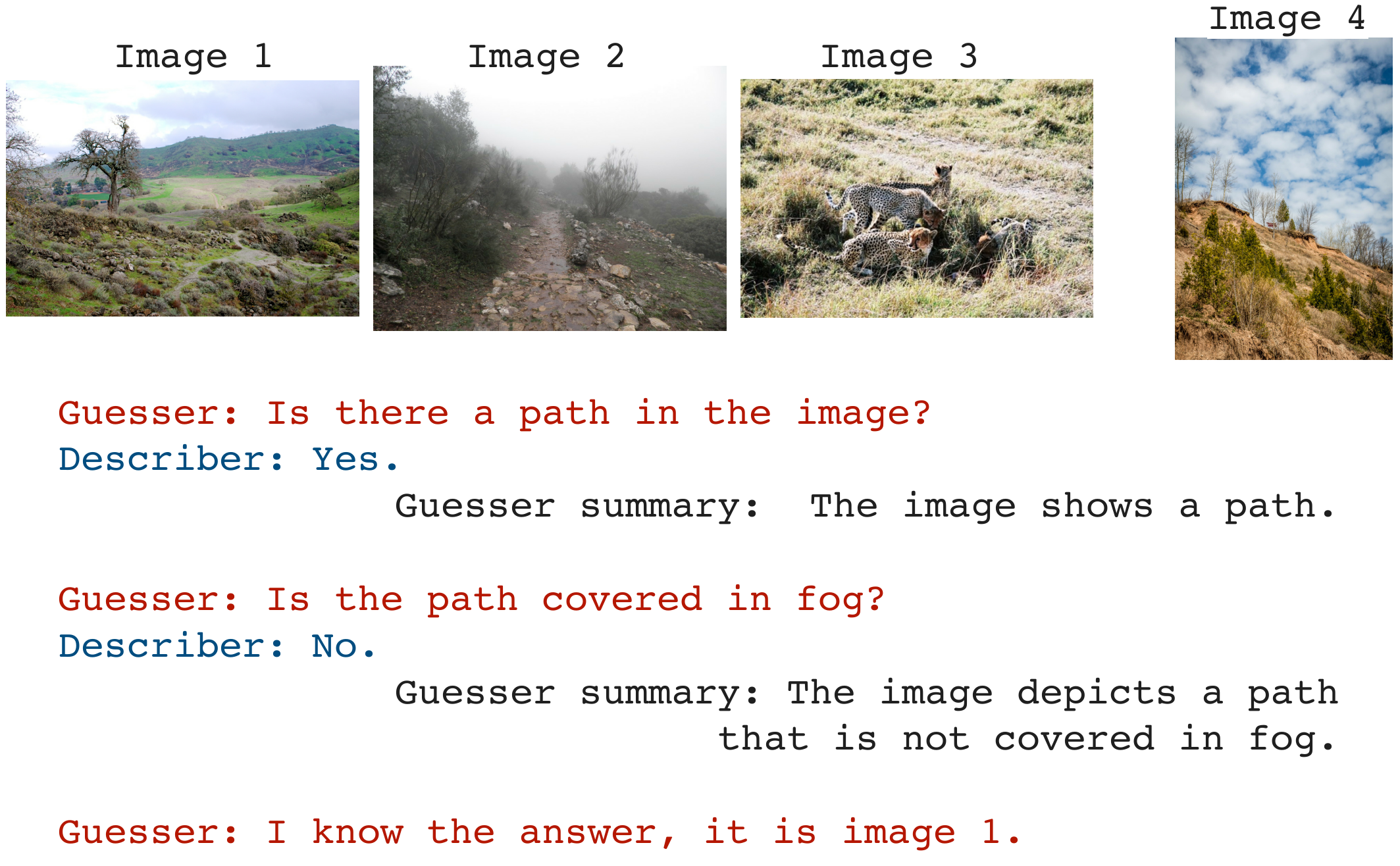}
    \caption{\textbf{An example of a dialog game with OpenImages} grouped by the image similarity.
    The figure shows the Guesser's questions, the Describer's answers, and the Guesser's internal dialog summary.  The Guesser correctly identifies the target image (1) at the end of the dialog.}
    \vspace{-3mm}
    \label{fig:open_image_example_dialog}
\end{figure}

\paragraph{OpenImages}
We select a subset of \num{1000} random images, forming them into games with $N=4$ images.
As the dataset does not contain clusters, we select the most similar images \citep{jia2021align} as distractors.
An example of a dialog game produced in this scenario is demonstrated in Figure~\ref{fig:open_image_example_dialog}.

\subsubsection{Evaluations Tasks}
\paragraph{Dialog success rate}

Following prior work using dialog games to assess VLM capabilities \citep{hakimov2024usinggameplayinvestigate}, we use the dialog game success rate as one of measures of the model's improvement.
We report the percentage of games where the Guesser correctly identifies the target image across all $N$ tested permutations (as described in Section~\ref{method-filtering}).  

\paragraph{Visual question answering (VQA)}

To assess the broader impact of our self-improvement method on general visual understanding, we evaluate the fine-tuned model on a subset of the VQAv2 dataset~\citep{goyal2017making}.
We focus on two specific question types:

\begin{itemize}
    \item \textbf{Binary (yes/no) questions}: Semantically equivalent phrasings (e.g., "No" and "There is no cat") are treated as correct. We report the model accuracy.
    \item \textbf{Object counting questions}: All answers and ground truth labels are converted to numerical form (e.g., "one" becomes "1", "none" becomes "0"). We report a strict exact-match accuracy.
\end{itemize}

\subsubsection{Results}

Table~\ref{tab:docci_captioning} compares the performance of the base Gemini 1.5 Flash model with VLM Dialog Games method.
Fist, results demonstrate that the VLM Dialog Games method with either the DOCCI or OpenImages datasets improves performance within the game with both training and unseen images (e.g., games played on DOCCI by a model trained with OpenImages).
More importantly we also achieve better performance on broader visual understanding tasks as measured by VQA accuracy.
Note that evaluation images for it are drawn from a distinct dataset (VQAv2), demonstrating the generalization of our method.
Specifically, for DOCCI dialog games, the accuracy on the VQAv2 yes/no and counting subsets increased by \num{6.8}\% and \num{2.3}\%, respectively.
For OpenImages dialog games, yes/no question accuracy increases by \num{10.4}\% and remains unchanged for counting questions. 
We hypothesis that different image sources may be better suited for improving specific tasks.
For example, \citet{OnoeDocci2024} note that many DOCCI images contain references to counts, suggesting that this dataset is well-suited for self-improvement on counting task.

\begin{table*}[h]
    \centering
     \caption{\textbf{Comparison of VLM Dialog Games and the initial Gemini 1.5 Flash.} Fine-tuning on dialog game data improves both game success rate and VQA performance (yes/no and counting subsets).  Results demonstrate generalization across training and evaluation datasets.}
    \vspace{5mm}
    \begin{tabular}{l|r|r|r|r}
      \multicolumn{1}{c}{Model} & \multicolumn{2}{c}{game success} & VQA  & VQA \\
       & ~~DOCCI~~ & OpenImages & yes/no & counting \\
      \midrule
      Gemini 1.5 Flash & 20.3\%  & 18.4\% & 73.0\% & 56\% \\
      VLM Dialog Games (DOCCI) & 24.4\% & 21.9\% & 79.8\% (+6.8) & 58.3\% (+2.3) \\
      VLM Dialog Games (OpenImages) & 25.6\% & 23.6\% & 83.4\% (+10.4) & 56\% (+0.0)\\
    \end{tabular}
    \label{tab:docci_captioning}
\end{table*}

\subsection{Ablation Studies}
\label{sec:exps-openimages-ablations}

Next, we investigate the impact of key design choices: the number of images per game and the method of image grouping.
We test the different options on OpenImages dialog games and VQA yes/no question accuracy.

\paragraph{Impact of the number of images per game}
We study the effect of $N$ on the game complexity by varying $N$ from \num{2} to \num{8} (see Appendix~\ref{game-examples-n} for dialog examples). Table~\ref{tab:openimage_n_images} presents the game success rate, the number of question-answer pairs from successful dialogs, and the VQAv2 yes/no accuracy for each $N$.
While fine-tuning with data from any $N$ improves VQAv2 performance compared to the base Gemini 1.5 Flash model, the best result is achieved with $N = 4$ in this study.
With $N = 2$, the game is relatively simple, leading to a high success rate but potentially less informative data, and a higher probability of erroneous data due to the correct guesses by chance.
Conversely, with $N = 8$, the game becomes too difficult, resulting in few successful dialogs for fine-tuning.
These results confirm that balancing game difficulty and the quantity of training data is crucial for generating an optimal dataset for fine-tuning.

\begin{table}[t]
    \centering
    \caption{\textbf{Impact of varying the number of images $N$ per game}: We report the number of successful dialog games (out of \num{1000}), the total number of question-answer pairs extracted, and the VQAv2 yes/no accuracy after fine-tuning. The optimal $N$ in this case appears to be \num{4}, balancing game difficulty and data quantity.}
    \vspace{5mm}
    \begin{tabular}{c|r|r|r}
      $N$ & game  & QA & VQA \\
       & success & pairs & yes/no \\
      \midrule
      2  & 83.7\% & 879 & 81.3\%  (~~+8.3\%) \\
      4  & 18.4\% & 275 & 83.4\% (+10.4\%) \\
      8  & 0.24\% & 34 & 77.1\%  (~~+4.1\%) \\
      \midrule
      \multicolumn{3}{l}{Gemini 1.5 Flash} & \multicolumn{1}{|l}{73\%} \\
    \end{tabular}
    \label{tab:openimage_n_images}
\end{table}

\paragraph{Impact of Image Grouping Strategy}

We investigate how image grouping affects model performance by comparing two strategies: 1) similarity-based grouping (Section~\ref{sec:game-config}), which uses visually and conceptually related distractors to elicit more targeted Guesser questions, and 2) random distractor selection.
Table~\ref{tab:openimage_vqav2_grouping} compares models using these strategies. 
Both strategies improve over the initial Gemini 1.5 Flash checkpoint ($73.0$\%) significantly, therefore, the VLM Dialog Game can be effectively implemented even with random image groupings. 
However, using similar images yields slightly higher accuracy ($83.4$\% vs. $82.6$\%).
While random images produce a larger quantity of successful dialogs ($24.7$\% vs. $18.4$\%), the increased challenge of similar images in a game likely leads to more informative training data.
Thus, we believe that for the best results in fine-tuning, we need to find a right trade off between game difficulty and training data quantity.

\begin{table}[t]
    \centering
    \caption{\textbf{Impact of image grouping strategy:} Both random and semantically similar image groupings lead to significant performance gains compared to the baseline. Although using semantically similar images demonstrates slightly better results, the difference is small, highlighting the robustness of the VLM Dialog Game approach even with random image selection.}
    \vspace{5mm}
    \begin{tabular}{l|r|r}
      Image grouping & game & VQA \\
      strategy & success & yes/no \\
      \midrule
      None (initial) & N/A & 73.0\% \\
      Similar images & 18.4\% & 83.4\% (+10.4\%) \\
      Random images & 24.7\% & 82.6\% (~~+9.6\%) \\
    \end{tabular}
    \label{tab:openimage_vqav2_grouping}
\end{table}

\subsection{Robotics Dialog Games}
\label{sec:exps-robotics}

\begin{figure}[t]
    \centering
    \includegraphics[width=\columnwidth]{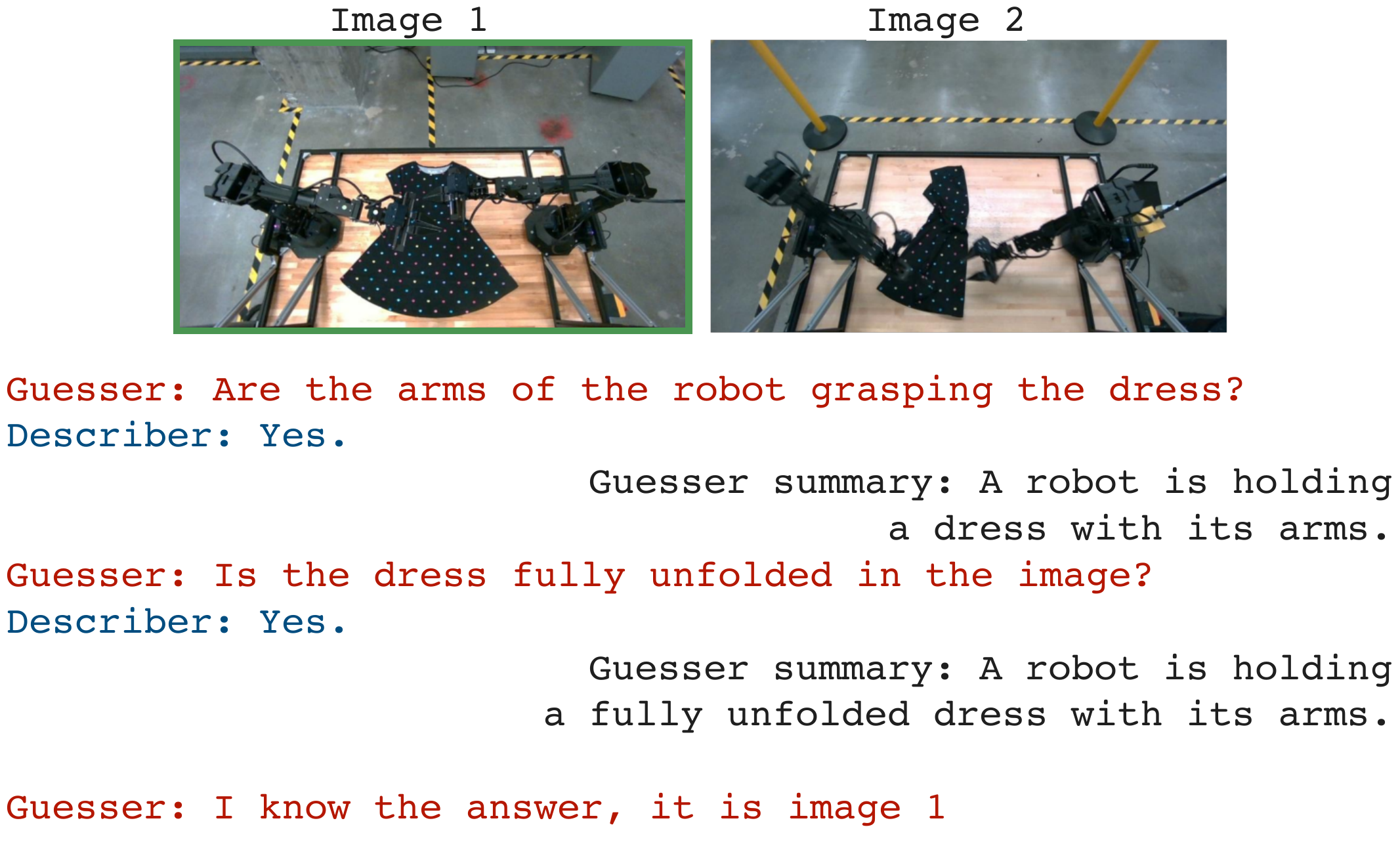}
    \caption{
    \textbf{An example of a dialog game in the robotics domain.} The figure shows the Guesser's questions, the Describer's answers, and the Guesser's internal dialog summary.  The Guesser correctly identifies the target image (1) at the end of the dialog.}
    \vspace{-3mm}
    \label{fig:game-example-robotics-success}
\end{figure}

High-quality interleaved data is scarce in specialized domains, potentially limiting base model performance in applications.  
This section describes our experiments using the VLM Dialog Games on video frames from a robotics manipulation domain where we test VLM success detection in object manipulation tasks.

\subsubsection{Dataset and Game Configuration}
\label{sec:exps-robotics-setup}

We use image frames from videos recorded in the ALOHA setup (A Low-cost Open-source Hardware System for Bimanual Teleoperation)~\citep{zhao2023learning}.
The images feature bimanual robotic arms performing \num{10} object manipulation tasks (e.g., putting objects in containers).
We use images captured from an overhead camera perspective. 
Our dataset comprises \num{20} episodes (both successful and unsuccessful) for each of the \num{10} tasks, totaling \num{200} episodes. 
We limit the game to only two images randomly sampled from the \textit{same task} execution as the success rate drops significantly with more images.
We generate \num{1000} games for each of the \num{10} tasks by sampling different frame combinations.
Figure \ref{fig:game-example-robotics-success} shows a dialog game example.

\subsubsection{Evaluation Task: Success Detection in Robotics}

To evaluate the impact of our method on robotic task understanding, we measure the model's ability to perform success detection. 
Accurate success detection is critical for various robotics applications, including policy training, evaluation, and data curating.
We evaluate success detection on the final frame of video episodes, treating it as a zero-shot VQA task~\citep{du23successvqa}. 
The model is presented with the final frame image and a textual description of the intended task (e.g., "open the drawer") and it is prompted with a question on task completion (e.g., "Is the drawer open?").
We report the accuracy of the model's yes/no responses.

\subsubsection{Baselines}
\label{sec:exps-robotics-baselines}

To isolate the specific contribution of the VLM Dialog Games, we compare our method against the original Gemini 1.5 Flash model and several other baselines.

\paragraph{Description Supervised Fine-Tuning (SFT-Description)}
Since our dialog games design utilizes task descriptions for each robotic episode, we include a baseline fine-tuned directly on image-description pairs.
This baseline "SFT-Description" helps determine if simply exposing the model to paired image and task descriptions from the target domain is sufficient for improvement.

\paragraph{Self-Improving Question Answering (Self-QA)}
This baseline explores an alternative self-improvement approach based on question answering similar to the approach of~\citet{luu2024questioning} (without the image captioning).
The model performs two tasks:
\begin{enumerate}
    \item \textbf{Question generation:} Given an image from the ALOHA dataset, the model generates a question about the scene.
    \item \textbf{Answer generation:} Given an image and a generated question, the model provides an answer.
\end{enumerate}

The prompts used for these tasks are detailed in Appendix~\ref{qa-prompts}.
This baseline tests whether a simpler self-improvement loop without the goal-oriented dialog structure can achieve similar results.

\paragraph{VLM Dialog Games (Answers Only)}
Our fine-tuning data includes both Describer and Guesser perspectives. 
Since the final success detection task closely resembles the Describer's role of answering questions, we include a baseline fine-tuned only on the datapoints from the Describer.
This isolates the contribution of the Guesser's questions to the overall improvement.

\begin{table*}[t]
    \centering
    \caption{\textbf{Success detection accuracy on the ALOHA dataset}, averaged across \num{10} tasks.  Fine-tuning on dialog game data outperforms the initial checkpoint and the other baselines. Iterative refinement further improves performance.}
    \vspace{5mm}
    \begin{tabular}{l|r|r}
      Model   & Game Success & Success Detection Accuracy \\
      \midrule
      Gemini 1.5 Flash & 14.39\% & 56.5\% \\
      VLM Dialog Games (round 1) & 40.15\% (+25.76\%) & 69.5\% (+13.0\%) \\
      VLM Dialog Games (round 2) & 53.74\% (+39.35\%) & 73.0\% (+16.5\%) \\
      \midrule
      SFT-Description & N/A & 65.0\% (~~+8.5\%) \\
      Self-QA & N/A & 67.0\% (+10.5\%) \\
      VLM Dialog Games (answers only) & 17.92\% (+3.53\%) & 68\% (+12.5)\% \\
    \end{tabular}
    \label{tab:robotics_result}
\end{table*}

\paragraph{Multiple Rounds of Self-Improvement}
We expect fine-tuning to improve the model's performance in subsequent games.
Thus, we use the improved model to generates a new, higher-quality dataset of synthetic dialogs. 
These dialogs are filtered and used to fine-tune the next iteration of the model, a process we refer to as "round 1" and "round 2".

In all cases we generate datasets with a size equivalent to the corresponding dialog game dataset and use it to fine-tune the Gemini 1.5 Flash model with the same settings.

\subsubsection{Results}
\label{sec:exps-robotics-results}

Table~\ref{tab:robotics_result} presents the success detection accuracy and game success rates averaged across the $10$ robotic tasks.
The initial Gemini 1.5 Flash model achieves a success detection accuracy of $56.5$\% on this highly specialised domain, only slightly above chance. 
Both the SFT-Description and Self-QA baselines improve upon this, demonstrating the benefit of domain-specific fine-tuning ($65.0$\% and $67.0$\% accuracy, respectively).

However, fine-tuning on a single round of dialog game data (VLM Dialog Games (round 1)) yields a larger improvement, achieving a success detection accuracy of $69.5$\% surpassing the baseline Self-QA by $2.5$\%.
Interestingly, although the VLM received no explicit instructions for success detection, the need to distinguish between frames from the \textit{same} task type lead it to focus on the task progression.
In contrast, the Self-QA method primarily generated object-related questions (see Appendix~\ref{sec:qa-examples} for examples).

Importantly, this initial round of dialog game fine-tuning also substantially increases the game success rate, from $14.39$\% to $40.15$\%, thus enabling further improvement.
We performed a second round of fine-tuning (VLM Dialog Games (round 2)), using data generated by the round 1 model.
This further boosted both the game success rate (to $53.74$\%) and the success detection accuracy (to $73.0$\%), a $16.5$\% absolute improvement over the original base model.

The VLM Dialog Games (answers only) baseline, which uses only the Describer's answers from the dialog games, achieves a success detection accuracy comparable to VLM Dialog Games (round 1). 
However, its game success rate remains comparatively low ($17.92$\%) and does not enable further iterative improvement.
This suggests that while the Describer's answers are sufficient for improving success detection, the Guesser's questions play a crucial role in improving the model's ability to play the dialog game effectively, which is necessary for continued self-improvement.

To conclude, our dialog game framework enables significant adaptation to specialized tasks like robotic success detection, where standard VLM pre-training may be less effective due to the lack of the domain-specific data.
Crucially, this self-improvement is achieved with minimal task-specific supervision, requiring only video episodes to guide the dialog generation.

\section{Discussion, limitations and conclusion}
\label{sec:conclusions}

This paper introduced VML Dialog Games as a novel self-improvement framework.
Our approach leverages goal-oriented self-play between two agents engaged in a reference-style dialog game. 
By automatically filtering for successful game interactions, we generate a high-quality dataset of interleaved image and text data.
We demonstrated, through experiments on both general visual question answering (using OpenImages, DOCCI and VQAv2) and robotic success detection (using ALOHA), that fine-tuning VLMs on this synthetically generated data leads to significant performance improvements. 
Crucially, our approach requires minimal supervision, demonstrating the potential for scalable and data-efficient VLM training.

Despite promising results, our approach has limitations.
First, effectiveness depends on the initial VLM's instruction-following, as the game can only refine the existing capabilities.
Second, agents might discover trivial or useless "winning" strategies (e.g., querying specific pixel colors or inventing a private vocabulary) without genuine understanding.
While we saw significant improvement in robotics (likely due to its under-representation in pre-training), iterative gains plateaued quickly, and scalability requires further investigation. Finally, some tasks may be solvable via prompt engineering and inference-time computation, potentially avoiding fine-tuning costs.

We believe that the success of our method suggests a promising direction for model development.
While current VLMs continue to improve, the core principle of our approach – learning from successful interactions in a goal-oriented setting – remains applicable. 
Multimodal dialog games offer a general recipe for VLM self-improvement, adaptable to various domains and tasks, particularly those with scarce or specialized data. 
Future work could explore different game designs, prompting strategies, and methods for identifying and utilizing dialog interactions.

\section*{Acknowledgements}
We would like to thank Todor Davchev, Thomas Lampe, Murilo Martins and Jingwei Zhang as well as Ira Ktena, Ryutaro Tanno and David Barrett for participating in the project hackathons.
Also, we are thankful to Junkyung Kim and Jason Baldridge for thoughtful feedback on the manuscript.

\newpage
\bibliography{main}

\begin{thebibliography}{26}
\providecommand{\natexlab}[1]{#1}
\providecommand{\url}[1]{\texttt{#1}}
\expandafter\ifx\csname urlstyle\endcsname\relax
  \providecommand{\doi}[1]{doi: #1}\else
  \providecommand{\doi}{doi: \begingroup \urlstyle{rm}\Url}\fi

\bibitem[Bai et~al.(2022)Bai, Kadavath, Kundu, Askell, Kernion, Jones, Chen,
  Goldie, Mirhoseini, McKinnon, et~al.]{bai2022constitutional}
Y.~Bai, S.~Kadavath, S.~Kundu, A.~Askell, J.~Kernion, A.~Jones, A.~Chen,
  A.~Goldie, A.~Mirhoseini, C.~McKinnon, et~al.
\newblock Constitutional {AI}: Harmlessness from {AI} feedback.
\newblock \emph{arXiv:2212.08073}, 2022.

\bibitem[Chalamalasetti et~al.(2023)Chalamalasetti, G{\"o}tze, Hakimov,
  Madureira, Sadler, and Schlangen]{chalamalasetti2023clembench}
K.~Chalamalasetti, J.~G{\"o}tze, S.~Hakimov, B.~Madureira, P.~Sadler, and
  D.~Schlangen.
\newblock clembench: Using game play to evaluate chat-optimized language models
  as conversational agents.
\newblock In \emph{EMNLP}, 2023.

\bibitem[Chen et~al.(2024)Chen, Deng, Yuan, Ji, and Gu]{chen2024selfplay}
Z.~Chen, Y.~Deng, H.~Yuan, K.~Ji, and Q.~Gu.
\newblock Self-play fine-tuning converts weak language models to strong
  language models.
\newblock In \emph{ICML}, 2024.

\bibitem[Das et~al.(2017)Das, Kottur, Gupta, Singh, Yadav, Moura, Parikh, and
  Batra]{das2017visual}
A.~Das, S.~Kottur, K.~Gupta, A.~Singh, D.~Yadav, J.~M. Moura, D.~Parikh, and
  D.~Batra.
\newblock Visual dialog.
\newblock In \emph{CVPR}, 2017.

\bibitem[De~Vries et~al.(2017)De~Vries, Strub, Chandar, Pietquin, Larochelle,
  and Courville]{de2017guesswhat}
H.~De~Vries, F.~Strub, S.~Chandar, O.~Pietquin, H.~Larochelle, and
  A.~Courville.
\newblock Guesswhat?! visual object discovery through multi-modal dialogue.
\newblock In \emph{CVPR}, 2017.

\bibitem[Du et~al.(2023)Du, Konyushkova, Denil, Raju, Landon, Hill, de~Freitas,
  and Cabi]{du23successvqa}
Y.~Du, K.~Konyushkova, M.~Denil, A.~Raju, J.~Landon, F.~Hill, N.~de~Freitas,
  and S.~Cabi.
\newblock Vision-language models as success detectors.
\newblock In \emph{Proceedings of The 2nd Conference on Lifelong Learning
  Agents}, 2023.

\bibitem[Gemini(2024)]{gemini2024}
T.~Gemini.
\newblock Gemini 1.5: {U}nlocking multimodal understanding across millions of
  tokens of context.
\newblock \emph{arXiv:2403.05530}, 2024.

\bibitem[Goyal et~al.(2017)Goyal, Khot, Summers-Stay, Batra, and
  Parikh]{goyal2017making}
Y.~Goyal, T.~Khot, D.~Summers-Stay, D.~Batra, and D.~Parikh.
\newblock Making the {V} in {VQA} matter: Elevating the role of image
  understanding in visual question answering.
\newblock In \emph{CVPR}, 2017.

\bibitem[Haber et~al.(2019)Haber, Baumg{\"a}rtner, Takmaz, Gelderloos, Bruni,
  and Fern{\'a}ndez]{haber2019photobook}
J.~Haber, T.~Baumg{\"a}rtner, E.~Takmaz, L.~Gelderloos, E.~Bruni, and
  R.~Fern{\'a}ndez.
\newblock The {P}hoto{B}ook dataset: Building common ground through
  visually-grounded dialogue.
\newblock In \emph{ACL}, 2019.

\bibitem[Hakimov et~al.(2024)Hakimov, Abdullayeva, Koshti, Schmidt, Weiser,
  Beyer, and Schlangen]{hakimov2024usinggameplayinvestigate}
S.~Hakimov, Y.~Abdullayeva, K.~Koshti, A.~Schmidt, Y.~Weiser, A.~Beyer, and
  D.~Schlangen.
\newblock Using game play to investigate multimodal and conversational
  grounding in large multimodal models.
\newblock \emph{arXiv:2406.14035}, 2024.

\bibitem[Huang et~al.(2023)Huang, Gu, Hou, Wu, Wang, Yu, and
  Han]{huang2022large}
J.~Huang, S.~S. Gu, L.~Hou, Y.~Wu, X.~Wang, H.~Yu, and J.~Han.
\newblock Large language models can self-improve.
\newblock In \emph{EMNLP}, 2023.

\bibitem[Jia et~al.(2021)Jia, Yang, Xia, Chen, Parekh, Pham, Le, Sung, Li, and
  Duerig]{jia2021align}
C.~Jia, Y.~Yang, Y.~Xia, Y.-T. Chen, Z.~Parekh, H.~Pham, Q.~Le, Y.-H. Sung,
  Z.~Li, and T.~Duerig.
\newblock Scaling up visual and vision-language representation learning with
  noisy text supervision.
\newblock In \emph{ICML}, 2021.

\bibitem[Kaplan et~al.(2020)Kaplan, McCandlish, Henighan, Brown, Chess, Child,
  Gray, Radford, Wu, and Amodei]{kaplan2020scaling}
J.~Kaplan, S.~McCandlish, T.~Henighan, T.~B. Brown, B.~Chess, R.~Child,
  S.~Gray, A.~Radford, J.~Wu, and D.~Amodei.
\newblock Scaling laws for neural language models.
\newblock \emph{arXiv:2001.08361}, 2020.

\bibitem[Krauss and Weinheimer(1964)]{krauss1964changes}
R.~Krauss and S.~Weinheimer.
\newblock Changes in reference phrases as a function of frequency of usage in
  social interactions.
\newblock \emph{Psychonomic Science}, 1964.

\bibitem[Kuznetsova et~al.(2020)Kuznetsova, Rom, Alldrin, Uijlings, Krasin,
  Pont-Tuset, Kamali, Popov, Malloci, Kolesnikov, Duerig, and
  Ferrari]{kuznetsova2020openimages}
A.~Kuznetsova, H.~Rom, N.~Alldrin, J.~Uijlings, I.~Krasin, J.~Pont-Tuset,
  S.~Kamali, S.~Popov, M.~Malloci, A.~Kolesnikov, T.~Duerig, and V.~Ferrari.
\newblock The open images dataset v4: {U}nified image classification, object
  detection, and visual relationship detection at scale.
\newblock \emph{International journal of computer vision}, 2020.

\bibitem[Li et~al.(2023{\natexlab{a}})Li, Gu, Koner, Sharifzadeh, and
  Tresp]{li2023dalle}
H.~Li, J.~Gu, R.~Koner, S.~Sharifzadeh, and V.~Tresp.
\newblock Do {DALL-E} and {F}lamingo understand each other?
\newblock In \emph{ICCV}, 2023{\natexlab{a}}.

\bibitem[Li et~al.(2023{\natexlab{b}})Li, Bhardwaj, Tian, Zhang, Barber,
  Katabi, Lajoie, Chang, and Krishnan]{li2023leveraging}
T.~Li, S.~Bhardwaj, Y.~Tian, H.~Zhang, J.~Barber, D.~Katabi, G.~Lajoie,
  H.~Chang, and D.~Krishnan.
\newblock Leveraging unpaired data for vision-language generative models via
  cycle consistency.
\newblock \emph{arXiv 2310.03734}, 2023{\natexlab{b}}.

\bibitem[Luu et~al.(2024)Luu, Le, and Vo]{luu2024questioning}
D.-T. Luu, V.-T. Le, and D.~M. Vo.
\newblock Questioning, answering, and captioning for zero-shot detailed image
  caption.
\newblock In \emph{Proceedings of the Asian Conference on Computer Vision},
  2024.

\bibitem[Onoe et~al.(2024)Onoe, Rane, Berger, Bitton, Cho, Garg, Ku, Parekh,
  Pont-Tuset, Tanzer, Wang, and Baldridge]{OnoeDocci2024}
Y.~Onoe, S.~Rane, Z.~Berger, Y.~Bitton, J.~Cho, R.~Garg, A.~Ku, Z.~Parekh,
  J.~Pont-Tuset, G.~Tanzer, S.~Wang, and J.~Baldridge.
\newblock {DOCCI: Descriptions of Connected and Contrasting Images}.
\newblock In \emph{ECCV}, 2024.

\bibitem[OpenAI et~al.(2023)OpenAI, Achiam, Adler, Agarwal, Ahmad, Akkaya,
  Aleman, Almeida, Altenschmidt, Altman, Anadkat, et~al.]{achiam2023gpt}
OpenAI, J.~Achiam, S.~Adler, S.~Agarwal, L.~Ahmad, I.~Akkaya, F.~L. Aleman,
  D.~Almeida, J.~Altenschmidt, S.~Altman, S.~Anadkat, et~al.
\newblock Gpt-4 technical report.
\newblock \emph{arXiv:2303.08774}, 2023.

\bibitem[Sharifzadeh et~al.(2024)Sharifzadeh, Kaplanis, Pathak, Kumaran, Ilic,
  Mitrovic, Blundell, and Banino]{sharifzadeh2024synth}
S.~Sharifzadeh, C.~Kaplanis, S.~Pathak, D.~Kumaran, A.~Ilic, J.~Mitrovic,
  C.~Blundell, and A.~Banino.
\newblock Synth$^2$: Boosting visual-language models with synthetic captions
  and image embeddings.
\newblock \emph{arXiv:2403.07750}, 2024.

\bibitem[Subramaniam et~al.(2025)Subramaniam, Du, Tenenbaum, Torralba, Li, and
  Mordatch]{subramaniam2025multiagent}
V.~Subramaniam, Y.~Du, J.~B. Tenenbaum, A.~Torralba, S.~Li, and I.~Mordatch.
\newblock Multiagent finetuning: Self improvement with diverse reasoning
  chains.
\newblock \emph{arXiv:2501.05707}, 2025.

\bibitem[Wang et~al.(2024)Wang, Wang, Yu, Lu, Xiao, Sun, Liu, Zou, Gao, Yang,
  et~al.]{wang2024self}
J.~Wang, K.~Wang, Y.~Yu, Y.~Lu, W.~Xiao, Z.~Sun, F.~Liu, Z.~Zou, Y.~Gao,
  L.~Yang, et~al.
\newblock Self-improving generative foundation model for synthetic medical
  image generation and clinical applications.
\newblock \emph{Nature Medicine}, 2024.

\bibitem[Yuan et~al.(2024)Yuan, Pang, Cho, Li, Sukhbaatar, Xu, and
  Weston]{yuan2024selfrewarding}
W.~Yuan, R.~Y. Pang, K.~Cho, X.~Li, S.~Sukhbaatar, J.~Xu, and J.~E. Weston.
\newblock Self-rewarding language models.
\newblock In \emph{ICML}, 2024.

\bibitem[Zhao et~al.(2023)Zhao, Kumar, Levine, and Finn]{zhao2023learning}
T.~Z. Zhao, V.~Kumar, S.~Levine, and C.~Finn.
\newblock Learning fine-grained bimanual manipulation with low-cost hardware.
\newblock In \emph{RSS}, 2023.

\bibitem[Zhu et~al.(2017)Zhu, Park, Isola, and Efros]{zhu2017cyclegan}
J.-Y. Zhu, T.~Park, P.~Isola, and A.~A. Efros.
\newblock Unpaired image-to-image translation using cycle-consistent
  adversarial networks.
\newblock In \emph{ICCV}, 2017.

\end{thebibliography}

\section{Appendix: VLM Prompts}
\label{prompts}

\subsection{Prompt for VML Dialog Games}
\label{game-prompts}

Prompt to Describer agent to answer questions about the image faithfully is the same for all datasets and domains:

\begin{lstlisting}
You are given an image and your task is to answer a given question about it. Be precise and accurate. Only answer the question, do not say anything else about the image.
Image: {image}
Question: {question}
Answer:
\end{lstlisting}

Prompt for Guesser for general images:

\begin{lstlisting}
You are given several images Image 1, Image 2, Image 3, Image 4 and image description.
This image description refers to only a single image, however, the image description might be incomplete.
You task is the following:
1) If the image description can only refer to a single image from the set of images (Image 1, Image 2, Image 3, Image 4) you should provide the answer in the format:
Answer: I know the answer, it is image X.
where X is the index of an image (1, 2, 3, 4).
Only provide a response in this format when you are absolutely certain to which image the image description refers to.
Never provide an answer in this format when the image description is empty.
2) If no image description is provided or the image description can refer to more than one image, your task is to ask additional question to narrow down the space of possible images from the set (Image 1, Image 2, Image 3, Image 4).
Ask any question that would help you to narrow the space of possible images.
Choose a question that would help you to maximise the information about the content of the target image.
Try to ask objective, factual questions that cover the content of the image, but not the deductions about the scene or any impressions about the image.
Follow the format:
Question: put your question here.
So, now given the image descriptions and 4 images, decide if you are going to make a guess (in that case produce an Answer) or ask a question (in that case produce a Question).
Image description: {image_description}
Image 1: {image1} Image 2: {image2} Image 3: {image3} Image 4: {image4}
\end{lstlisting}

In robotics experiments, the Guesser agent was prompted with instructions very similar to the general images, but mentioning that the images come from a robotics domain.
These instructions also direct the agents to focus on relevant visual features for robotic manipulation tasks as opposed to details in the background (e.g., people who are passing by, chairs moved):

\begin{lstlisting}
You are given two images (Image 1, Image 2) from a scene where robot is trying to {task} and image description.
This image description refers to only a single image, however, the image description might be incomplete.
You task is the following:
1) If the image description can only refer to a single image from the set of images (Image 1, Image 2) you should provide the answer in the format:
Answer: I know the answer, it is image X.
where X is the index of an image (1,2).
Only provide an response in this format when you are absolutely certain to which image the image description refers.
Never provide an answer in this format when the image description is empty.
2) If no image description is provided or the image description can refer to more than one image, your task is to ask additional question to narrow down the space of possible images from the set (Image 1, Image 2).
Try to ask objective, factual questions that cover the content of the image.
Choose a question that would help you to maximise the information about the content of the image.
NEVER ask questions about the background of the robotic scene (e.g., people in the background, scooters or chairs).
NEVER ask questions about the facts that are already known from the image description.
Follow the format:
Question: put your question here

So, now given the image descriptions and 2 images, decide if you are going to make a guess (in that case produce an Answer) or ask a question (in that case produce a Question).
Image description: {image_description}
Image 1: {image1} Image 2: {image2}
\end{lstlisting}

Prompt to Guesser agent to summarise the state of the existing dialog for all domains:

\begin{lstlisting}
You are given a short description of a scene and one question and answer about it.
Your task is to summarise the content of the scene in a short sentence or paragraph. Only provide a summary, do no output anything else.
Always include all the details 1) from the description, 2) from question-answer pair into your summary.
Description: {description}
Question: {question}
Answer: {answer}
Your summary: 
\end{lstlisting}

\subsection{Full Prompts for the Baseline Self-QA}
\label{qa-prompts}

To ask questions:

\begin{lstlisting}
You are given an image and your task is to ask a question about the content of this image.
Try to ask objective, factual questions that cover the content of the image, but not the deductions about the scene or any impressions about the image.
NEVER ask questions about the background of the robotic scene (e.g., people in the background, scooters or chairs).
Follow the format:
Question: put your question here.
So, now given the image, ask a question.
Image: {image}
Question:
\end{lstlisting}

To answer the questions:
\begin{lstlisting}
You are given an image and your task is to answer a given question about it. Be precise and accurate. Only answer the question, do not say anything else about the image.
If possible, ONLY answer with *yes* or *no*.
Image: {image}
Question: {question}
Answer:
\end{lstlisting}

\section{Appendix: Dialog Examples for Varying Number of Images}
\label{game-examples-n}

\begin{figure}[t]
    \centering
    \includegraphics[width=1\columnwidth]{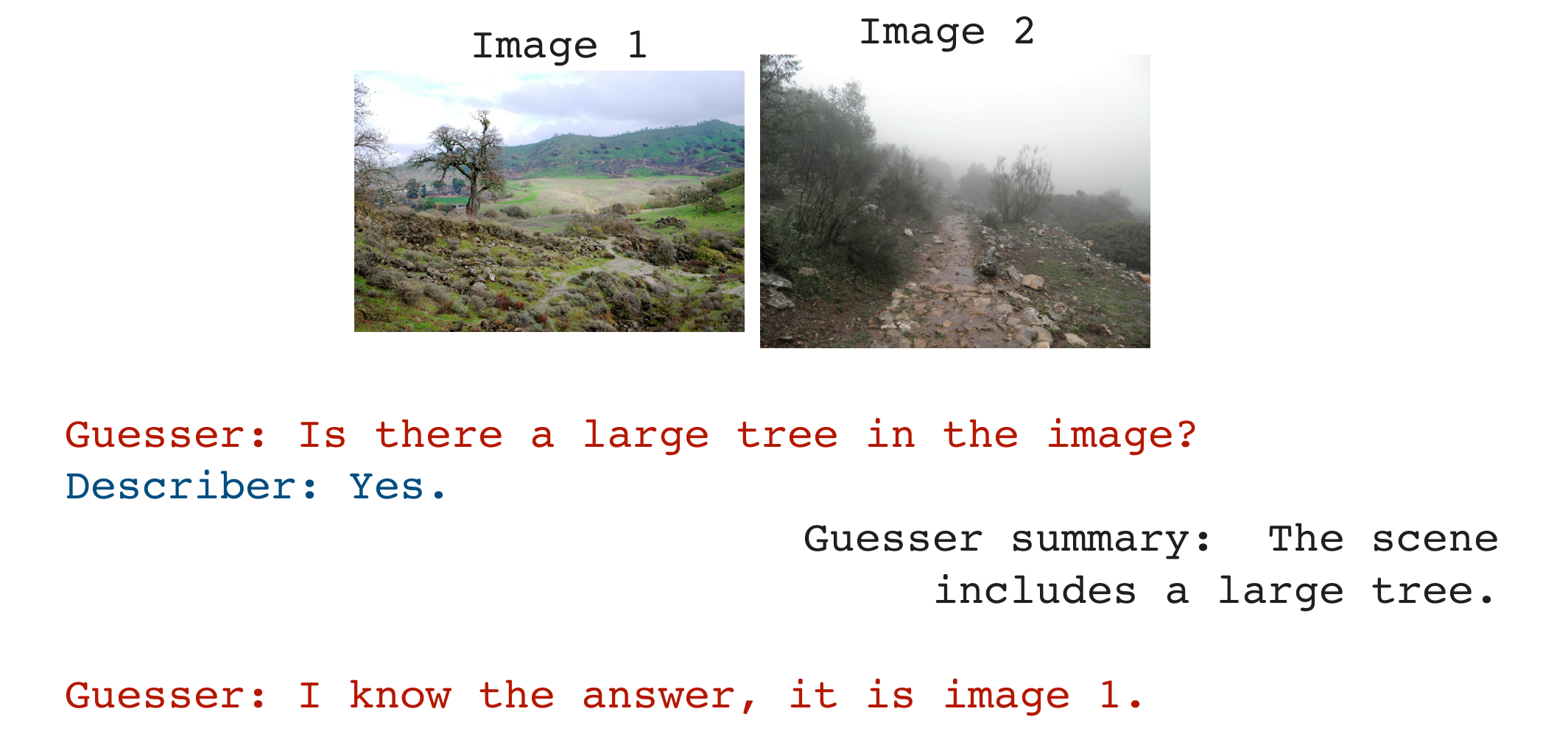}
    \caption{An example of a dialog game with two images.}
    \label{fig:dialog2}
\end{figure}

\begin{figure}[t]
    \centering
    \includegraphics[width=1\columnwidth]{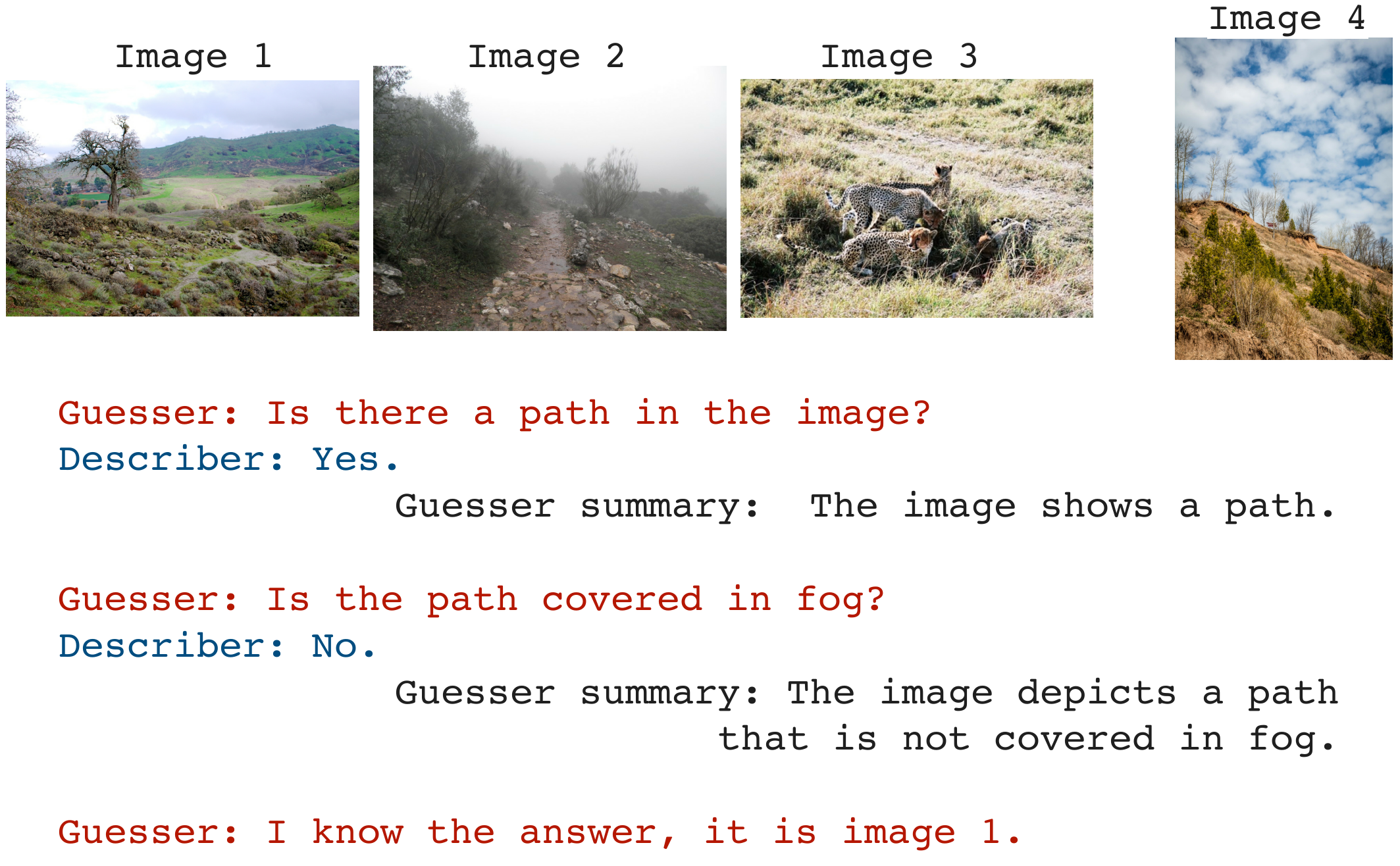}
    \caption{An example of a dialog game with four images.}
    \label{fig:dialog4}
\end{figure}

\begin{figure}[t]
    \centering
    \includegraphics[width=1\columnwidth]{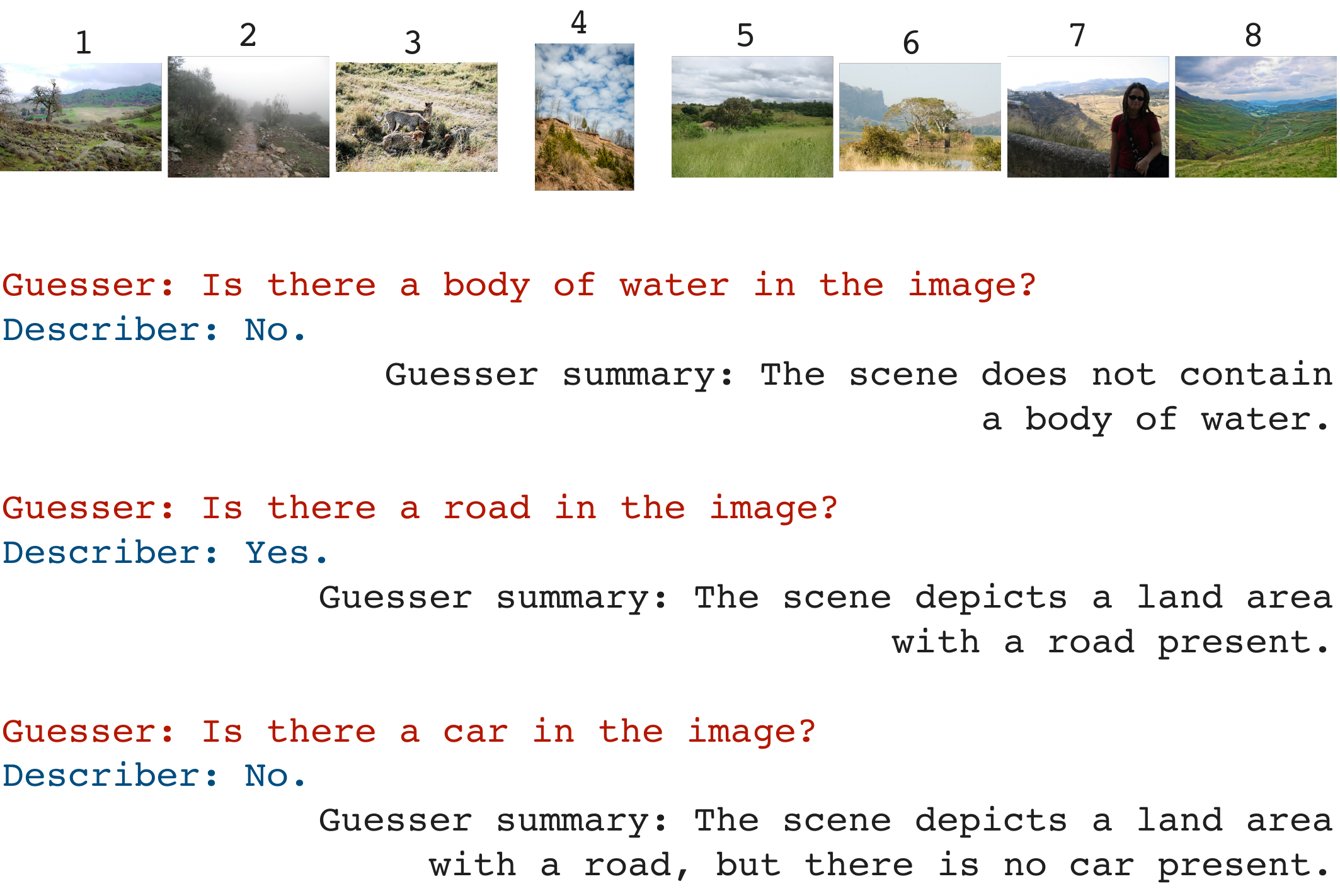}
    \caption{An example of a dialog game with eight images.}
    \label{fig:dialog8}
\end{figure}

Figures~\ref{fig:dialog2}, \ref{fig:dialog4}, and \ref{fig:dialog8} illustrate dialog examples with a consistent target image (Image 1) but a varying number of distractor images (1, 3, and 7, respectively, corresponding to $N$ values of 2, 4, and 8).
These examples demonstrate the effect of $N$ on dialog length and complexity.
With only two images ($N=2$), the dialog is short, focusing on a single distinguishing feature.
With four images ($N=4$), the dialog becomes more complex, requiring two questions that progressively narrow down the possibilities.
However, with eight images ($N=8$), the Guesser is unable to identify the target image within the three-question limit. 

\section{Appendix: Question-Answers Generated by Dialog Games and Self-QA}
\label{sec:qa-examples}

Self-QA:
\begin{itemize}
    \item Question: Is there a yellow object in the image? Answer: yes
    \item Question: Is there a red object on the surface? Answer: yes
    \item Question: Are there any balls inside the basket? Answer: No
    \item Question: Is there a red ball in the image? Answer: yes
    \item Question: Are there two robotic arms in the image? Answer: yes
    \item Question: Is there a red apple in the image? Answer: yes
    \item Question: Is there a red object on the floor? Answer: yes
    \item Question: Is there a piece of fruit in the basket? Answer: yes
    \item Question: Is there a yellow triangle in the image? Answer: yes
\end{itemize}

VML Dialog Games:
\begin{itemize}
    \item Question: Are there any Lego blocks on the floor that are not in the bag? Answer: Yes.
    \item Question: Is the drawer in the image open? Answer: Yes.
    \item Question: Is the trash bin lid open or closed? Answer: Closed.
    \item Question: Is the bowl inside the drying rack? Answer: No.
    \item Question: Is the cheese in the basket? Answer: No.
    \item Question: Is the banana inside the drying rack? Answer: Yes.
    \item Question: Is the robot's gripper holding the belt? Answer: No.
    \item Question: Is there a basket in the image? Answer: No.
    \item Question: Is the drawer open? Answer: Yes.
\end{itemize}

\section{Appendix: LLM Inference and Training Details}
\label{sec:training-details}

We rely on Gemini 1.5 Flash (gemini-1.5-flash-002) model which is available for inference and fine-tuning through the Google Cloud Vertex API.
For generating dailog and evaluation, we sample with nucleus sampling selecting the top \num{0.8} probability mass of tokens.
For the evalaution, we use sampling temperature \num{0}.
Our batch size for SFT is \num{16}, we use Adam optimizer with learning rate \num{5e-07}.
To prevent overfitting to the small datasets from the dialog games, we use a small unrelated to the any of the tested tasks dataset of images with text and track token loss on it.
We select a checkpoint just before the loss starts increasing. 
This usually corresponds to approximately one epoch of fine-tuning.

\newpage
\end{document}